\documentclass{article}

     \PassOptionsToPackage{numbers, compress}{natbib}

\usepackage[preprint]{neurips_2024}
\usepackage{amsmath}

\usepackage[utf8]{inputenc} 
\usepackage[T1]{fontenc}    
\usepackage{hyperref}       
\usepackage{url}            
\usepackage{booktabs}       
\usepackage{amsfonts}       
\usepackage{nicefrac}       
\usepackage{microtype}      
\usepackage{xcolor}         
\usepackage{graphicx}

\title{Evaluating geometric accuracy of NeRF reconstructions compared to SLAM method}

\author{%
  Adam Korycki \\
  Electrical and Computer Engineering\\
  UC Santa Cruz\\
  \texttt{akorycki@ucsc.edu}
  \And
  Colleen Josephson \\
  Electrical and Computer Engineering\\
  UC Santa Cruz\\
  \texttt{cojoseph@ucsc.edu}
 \And
  Steve McGuire \\
  Electrical and Computer Engineering\\
  UC Santa Cruz\\
  \texttt{stmcguir@ucsc.edu}
}

\begin{document}

\maketitle

\begin{abstract}

As Neural Radiance Field (NeRF) implementations become faster, more efficient and accurate, their applicability to real world mapping tasks becomes more accessible. Traditionally, 3D mapping, or scene reconstruction, has relied on expensive LiDAR sensing. Photogrammetry can perform image-based 3D reconstruction but is computationally expensive and requires extremely dense image representation to recover complex geometry and photorealism. NeRFs perform 3D scene reconstruction by training a neural network on sparse image and pose data, achieving superior results to photogrammetry with less input data. This paper presents an evaluation of two NeRF scene reconstructions for the purpose of estimating the diameter of a vertical PVC cylinder. One of these are trained on commodity iPhone data and the other is trained on robot-sourced imagery and poses. This neural-geometry is compared to state-of-the-art LiDAR-inertial SLAM in terms of scene noise and metric-accuracy.

\end{abstract}

\section{Introduction}

Three dimensional (3D) mapping is the task of digitally representing real-world scenes, typically in the form of a point cloud which discretely represents objects as dense sets of points in 3D space. These maps are of great importance to applications such as construction and city-planning which require the geometric properties of a given area. Another use of 3D reconstruction is in forest ecology, particularly for the task of mapping forests and estimating tree morphology. Tree diameters are used in global carbon accounting, a critical component for informed climate policy. Many cars have autonomous or semi-autonomous modes which rely on some form of 3D mapping to estimate the dynamic environment and ensure passenger safety. 

\par Traditionally, 3D reconstructions have relied on expensive light detection and ranging (LiDAR) sensing. LiDAR uses a spinning laser-beam deflection mechanism to measure the time-of-flight of light pulses and register the distances of surrounding scene geometry. A popular form of LiDAR mapping is a terrestrial laser scanning (TLS) system which is commonly used in construction and forestry. TLS involves a LiDAR mounted on a tri-pod and is used to generate single-scan reconstructions. This technique yields impressive results, but suffers from the difficulty of stitching many LiDAR scans together from different position, a necessary step to ensure accurate recovery of complex and occluded objects. Moreover, TLS systems typically cost 80,000-120,000 USD making this inaccessible to many users and applications. Research in using mobile robot platforms in combination with Simultaneous Localization and Mapping (SLAM) algorithms solves the single-scan problem using optimized pose-graphs to align thousands or millions of LiDAR scans taken along a robot's trajectory~\cite{mls, Freissmuth}. However, this solution comes with the cost of expensive 3D LiDAR and inertial measurement unit (IMU) hardware (10,000-25,000 USD).

\par Recent advancements in the computer vision and deep learning communities offer a new approach to 3D scene reconstruction. Most notably, Neural Radiance Fields (NeRFs)~\cite{nerf} provide the ability to recover intricate 3D geometry from a neural network trained on a sparse set of image views. This is not the first image-based mapping technique. Photogrammetry is as old as photography, and is the task of recovering 3D geometry from image data. The issue with traditional photogrammetry is the need for extremely dense image representation to recover complex detail. Moreover, this method is computationally expensive and struggles with representing complex scene artifacts such a lighting, glare, and shadowing. NeRFs are able to more accurately recover photo-realistic scene geometry with less input views. NeRFs do not require any special camera, and can be created from commodity mobile phone data. 

\par In this paper, we present an experiment which compares NeRF scene reconstruction with SLAM-based LiDAR-inertial mapping. The object of interest is a 40 cm diameter, three meter tall PVC pipe positioned in a concrete courtyard. Two NeRF reconstructions are created using two different data collection schemes. The first scheme uses an iPhone 14 camera with a free iOS application that provides camera poses. The second scheme involves robot-sourced imagery and camera poses derived from a LiDAR-inertial SLAM solution. These neural reconstructions are compared with the state-of-the-art method which uses a robot and LiDAR-inertial SLAM to create a dense point cloud representation of the PVC pipe. We compare the quality and noise of the maps, as well as the metric-relevance of the recovered PVC pipe diameter. The results of the experiment show that NeRF methods offer a promising avenue for accessible and and accurate 3D mapping without the need for expensive LiDAR sensors. 

\section{Related work}
\subsection{Simultaneous Localization and Mapping (SLAM)}
The SLAM problem can be broken into two tasks: building a map of the environment and estimating the robot's trajectory within that map, simultaneously. More specifically, given all sensor measurements $z_{1:T}$, all robot motion commands $u_{1:T}$, and an initial robot position $x_0$, estimate the posterior probability of the complete robot trajectory $x_{1:T}$ and the map \textbf{m} of the explored environment.
$$p(x_{1:T},\boldsymbol{m}|z_{1:T},u_{1:T},x_0)$$
This Bayesian problem formulation benefits from the ability to fuse several sensing modalities together while accounting for individual sensor noise. The modern-day solution to SLAM formulates the problem as a graph~\cite{slam} where each node represents a robot pose $x_i$ (relative 3D spatial location and orientation) and each edge represents a transformation $T_i$ between nodes. Edges also encapsulate loop-closure constraints. Loop-closure is the subprocess of identifying previously observed landmarks in order to correct for drift and error accumulation in the estimated trajectory. Based on the prior pose-graph, a global optimization is used to minimize the estimation error given the sensor measurements and loop-closure constraints. The resulting pose corrections are back propagated resulting in reduced error of scan alignment. 
\par SLAM has been extensively used to generate dense 3D scene reconstructions. In 2016, Pierzcha{\l}a et al.~\cite{mls} proposed using graph-SLAM in combination with a LiDAR-equipped UGV to map a forest inventory in southern Norway. The paper managed to achieve tree diameter estimation with -2.2 cm(2.3\%) RMSE over a 71 tree distribution using the RANSAC circle fitting method.

\subsection{Neural Radiance Fields (NeRFs)}
NeRF~\cite{nerf} is the current state-of-the-art solution to the problem of novel view synthesis. This problem involves generating an image of a 3D scene from a particular view when the only available information are images from other views. The two pivotal ingredients to the NeRF method are continuous volumetric representation and deep fully-connected network architecture. Their method inherits the exceptional photorealism and reconstruction fidelity of continuous representation and does so at a fraction of the storage cost compared to discrete approaches~\cite{nerf}. The scene is modeled as a multilayer perceptron network (MLP) which takes a 5D input vector composed of spatial location \textbf{x} = ($x, y, z$) and viewing angle \textbf{v} = ($\theta, \phi$) and learns the weights $\Theta$ to map each 5D coordinate to the corresponding 4D output vector of color \textbf{c} = (r, g, b) and volume density ($\sigma$). The 5D input space is sampled using ray tracing. The network architecture is two MLPs. The first is called the "coarse" network which casts the camera rays using stratified sampling, or in other words evenly distributed throughout the scene. This allows for a more informed sampling of input points along the camera rays. The second "fine" MLP is then trained using these optimized sampling locations. Lastly to create the full 3D reconstruction, volume rendering is used to minimize the error between the inferred views and ground truth training images. 

\section{Methods}
\subsection{LiDAR-inertial SLAM}
To represent a current state-of-the-art technique for 3D mapping, this paper considers LiDAR inertial odometry smoothing and mapping (LIOSAM). This method fuses LiDAR and IMU data together to create dense spatial reconstructions. LIOSAM uses the conventional pose-graph SLAM expression to optimize the generate map in real time. The platform used in this study is a Unitree B1 quadruped robot with a custom sensing payload. The LiDAR is an Ouster OS0-128 and the IMU is an Inertialsense IMX-5. LIOSAM runs onboard the robots computer which hosts a ROS framework running on Ubuntu 22.04. LIOSAM aligns the LiDAR frames together and on completion, provides a map of the explored environment and the trajectory of the robot.

\begin{figure}[h]
    \centering
    \includegraphics[width=3.5cm]{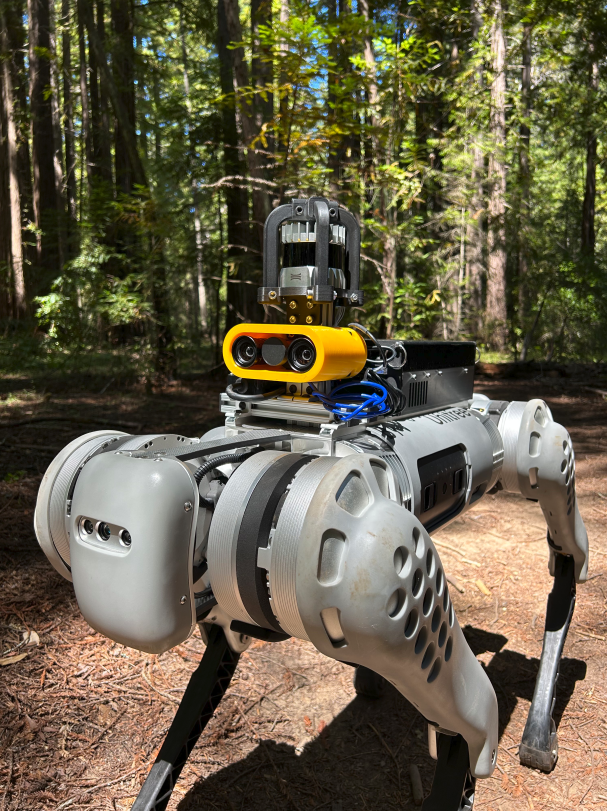}
    \centering
    \includegraphics[width=6cm]{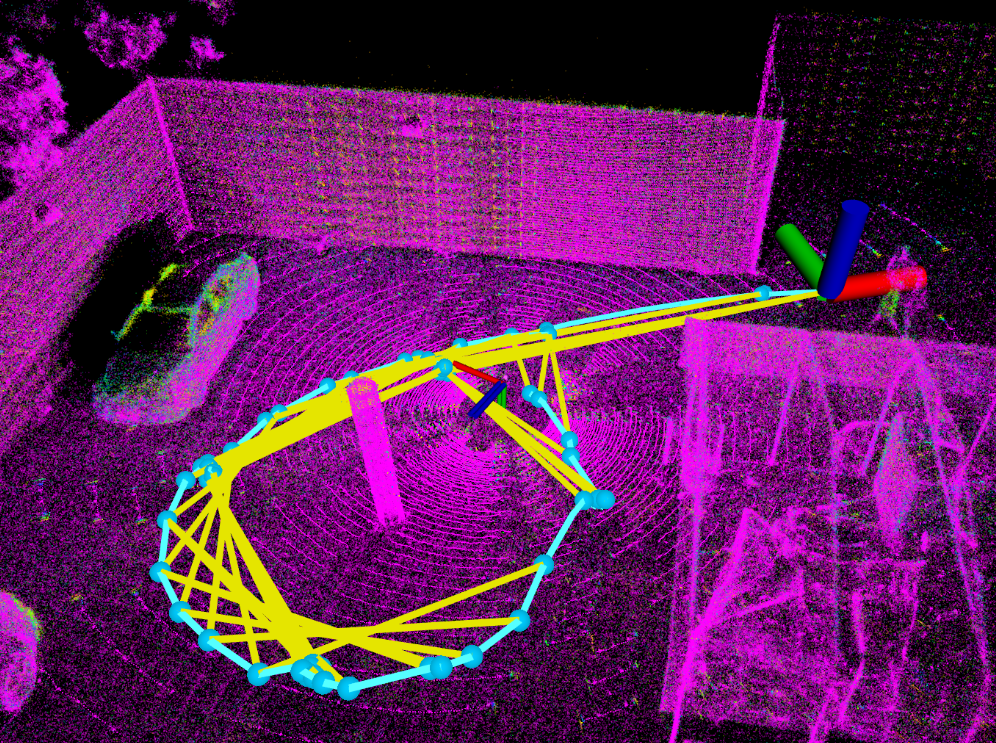}
    \caption{The robot platform (left) used for LiDAR-inertial SLAM reconstruction (right).}
    \label{fig:liosam}
\end{figure}

\subsection{NeRF reconstruction using nerfacto method}
Nerfacto is a method which draws from several published techniques~\cite{mip-nerf,nerf--,ref-nerf}, and has been shown to work very well on real data captured in a variety of environments. For this reason, Nerfacto was the chosen method for this thesis. Nerfacto improves on the base-NeRF method in a few key directions. The first of which is \textbf{pose refinement}. Error in image poses results in cloudy artifacts and loss of sharpness in the reconstructed scene. The Nerfacto method uses the back-propagated loss gradients to optimize the poses for each training iteration. Another improvement is in the ray-sampling of the 5D input space. Rays of light are modeled as conical frustums. A \textbf{piece-wise sampling} step uniformly samples the rays up to a certain distance from the camera origin, and then samples subsequent sections of the conical ray at step sizes which increase with each sample. This allows for high-detail sampling of close portions of the scene, while efficiently sampling distant objects as well. The output is fed into a \textbf{proposal sampler} which consolidates sample locations to sections of the scene which contribute most to the final 3D scene render. In order to inform which sample locations should be consolidated, serially connected density functions are used which are comprised of small fused-MLPs with hash-encoding. The output of these sampling stages is fed into the Nerfacto field. This stage incorporates \textbf{appearance embedding} which accounts for varying exposure among the training images. A "coarse" and "fine" pair of MLPs learn the output color and density based on the conditioned network input of spatial location (x, y, z) and viewing direction (dir).
\begin{figure}[h!]
    \centering
    \includegraphics[width=7cm]{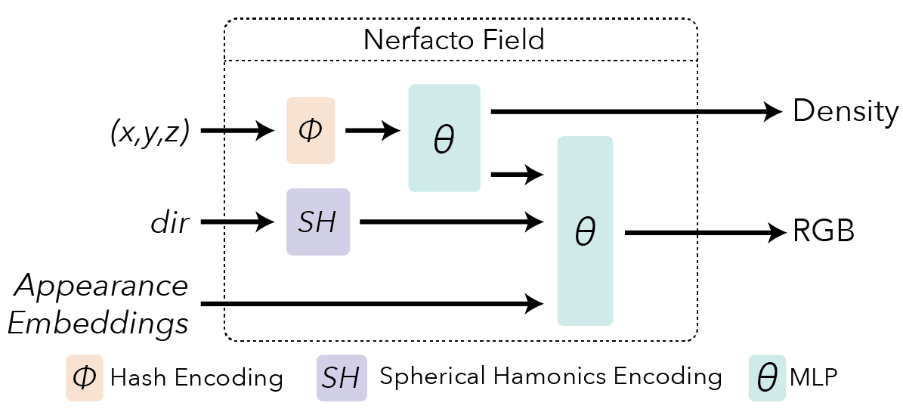}
    \caption{The Nerfacto network architecture comprised of two MLPs with positional encoding to approximate high-frequency volume functions and appearance embeddings to account for varying exposure in training images (Tancik et al.~\cite{nerfstudio}).}
    \label{fig:nerfacto}
\end{figure}
\par Posed image data is copied to a remote desktop PC for training. This computer hosts a 3.8 GHz AMD Ryzen Threadripper 3960X CPU with 24 cores, 64 GB DDR5 RAM, and 2 TB SSD storage. To support network training and rendering, the PC is also outfitted with two NVIDIA RTX-3070 graphics cards which aggregate to 16 GB of GDDR6 VRAM. The system runs Ubuntu 22.04 with CUDA-11.8 to interface with GPU hardware. Training a NeRF and exporting geometry with \textit{nerfstudio}~\cite{nerfstudio} is simple and straightforward.

\section{Experiment}
The experiment involved creating two NeRFs of a large PVC pipe in a courtyard. The first NeRF was trained on iPhone 14 images and poses provided by an iOS application called NeRFCapture~\cite{nerfcapture} which uses the ARKit toolbox to perform visual-inertial SLAM to provide camera poses. The second NeRF was trained on images taken by a quadruped robot. The poses were derived from LIOSAM state-estimation which provides the pose of the IMU. A rigid homogeneous transform is used convert the IMU pose to the camera frame. This transformation is derived through an offline visual-inertial calibration procedure. As a state-of-the-art comparison, a lidar-based point cloud is produced using the robot and the LIOSAM SLAM implementation. We evaluate the point cloud noise and the metric-accuracy of the PVC pipe reconstruction across the three methods.
\subsection{NeRF results from iPhone and robot imagery}
\begin{figure}[h!]
    \centering
    \includegraphics[width=8cm]{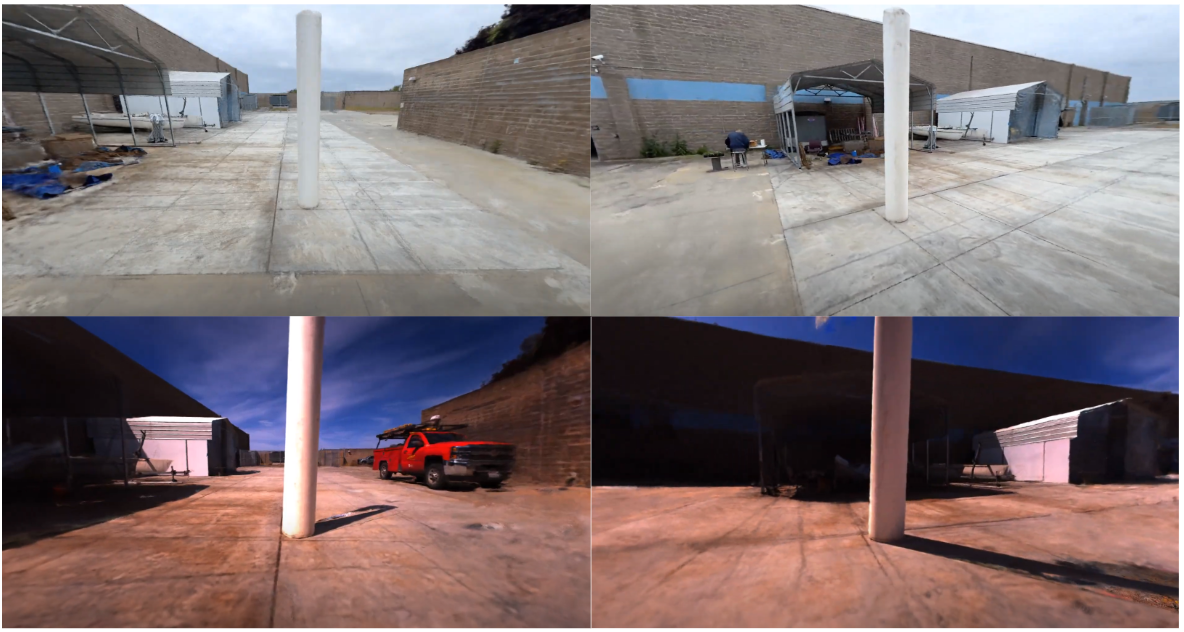}
    \caption{NeRF-synthesized novel views of the PVC pipe. Top views are neural renderings based on iPhone image and pose training data and bottom views are are rendered from robot sourced data.}
    \label{fig:pvc_pipe_nerfs}
\end{figure}

The rendered novel views in Figure~\ref{fig:pvc_pipe_nerfs} are visually impressive among both iPhone and robot NeRF methods. However, some holes are evident near shadows in the ground reconstruction of the robot NeRF which is likely due to harsh direct lighting. Table~\ref{table:pvc_pipe_results} shows that the iPhone NeRF outperformed the robot NeRF in both PSNR and LPIPS. Interestingly, the robot NeRF scored higher in SSIM. The overall greater performance of the iPhone-NeRFCapture method could be credited to its refined camera exposure controller, visually-informed pose estimation, and the Lambertian lighting conditions. The three generated point clouds were cropped using CloudCompare to center the PVC pipe in a 9 $m^2$ section of the ground. The two NeRF-sourced point clouds were able to be exported with an arbitrary number of points. However, this cropping reduced number of points in the LiDAR-inertial SLAM reconstruction by 95.5\% to 46,356 points. The massive decimation of the SLAM point cloud is due to the nature of points being registered by laser pulse returns which have a resolution of 262k points (128$\times$2048) per lidar frame. NeRFs do not face this issue since the geometry is created by sampling the learned color-ray space and filtering out low-density sections to only represent surfaces. This is a major advantage of NeRFs compared to lidar. 
\begin{table}[h!]
        \begin{center}
        \renewcommand{\arraystretch}{1.5}
            \begin{tabular}{ l|ccc|ccc } 
                Method & PSNR$\uparrow$ & SSIM$\uparrow$ & LPIPS$\downarrow$ & Resolution & Training set size & Total pixels\\ \hline
                NeRF-LIOSAM & 15.870 & \textbf{0.6749} & 0.4797 & 2448$\times$1700 & 186 & 774.1 M \\
                iPhone NeRF & \textbf{19.403} & 0.5494 & \textbf{0.4087} & 1920$\times$1440 & 103 & 284.7 M
            \end{tabular}
        \end{center}
    \caption{NeRF reconstruction metrics between iPhone and robot data collection methods. The iPhone NeRF outperformed the robot NeRF in both PSNR (higher is better) and LPIPS (lower is better). Interestingly, the robot NeRF scored better in SSIM (higher is better). The finer reconstruction quality from the iPhone could be attributed to better camera pose estimation, better exposure control, and Lambertian lighting.}
    \label{table:pvc_pipe_results}
\end{table}
\newpage
\subsection{Analyzing reconstruction fidelity and metric-accuracy}
\begin{figure}[h]
    \centering
    \includegraphics[width=10cm]{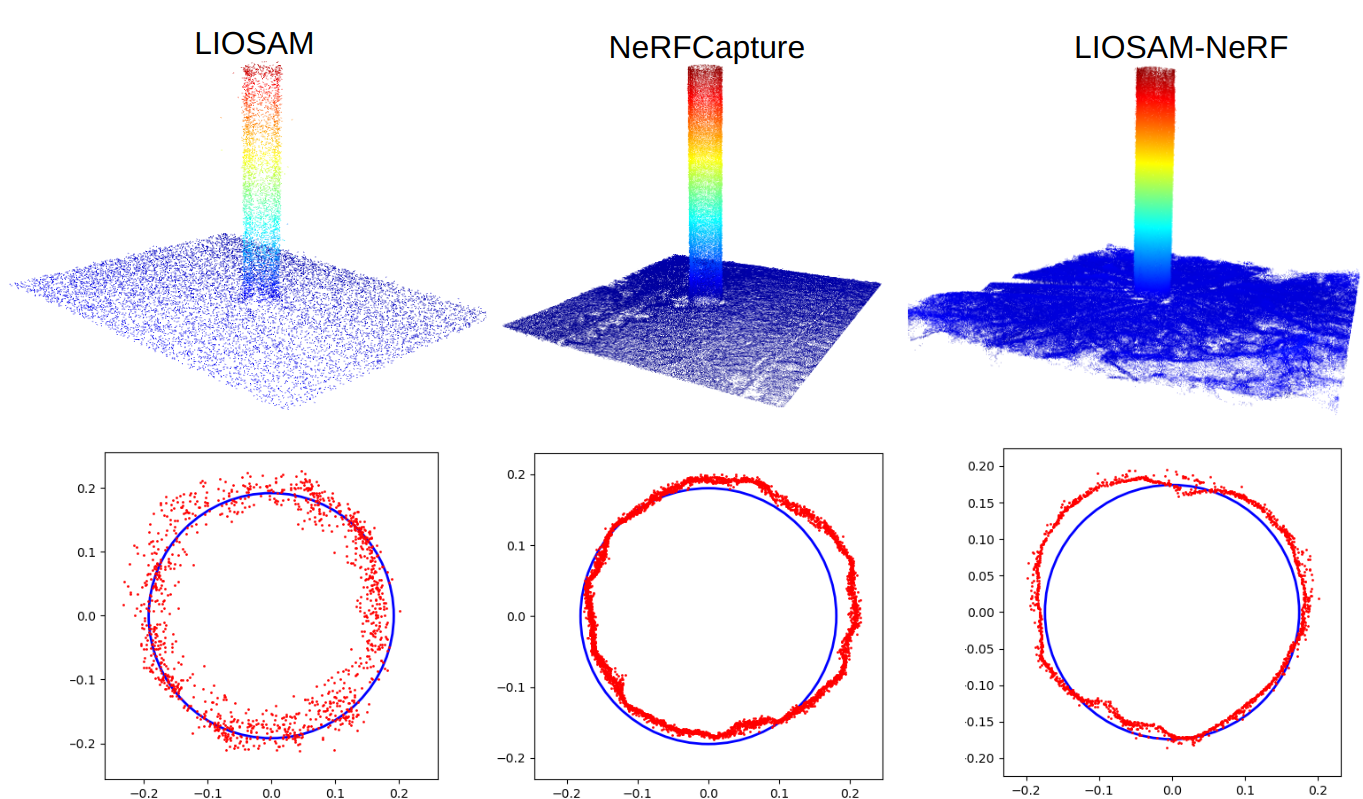}
    \caption{PVC pipe point cloud reconstructions produced by lidar-inertial SLAM (LIOSAM), NeRF-LIOSAM fusion, and NeRFCapture methods. Corresponding 2D projection of points used by TreeTool for cylinder-ellipse modeling are shown in red. The blue circles represent the fitted cylinder-ellipse models. These results provide two key insights. First, the RANSAC cylinder fitting method used in TreeTool is biased towards circular portions of the point cloud which explains the under-sizing trend in the estimated diameters. Second, the NeRF reconstructions have significantly less noise deviation compared to the SLAM reconstruction. This is likely due to sub-optimal LiDAR-inertial extrinsic calibration.}
    \label{fig:pvc_pipe_results1}
\end{figure}
TreeTool~\cite{treetool} was used to evaluate the diameter estimation accuracy between the three methods. The cropped point clouds and 2D projections of the points from each method are seen in Figure~\ref{fig:pvc_pipe_results1}. This figure provides several key insights. Most evidently, the deviation in the surface noise is much less in both NeRF methods compared to SLAM. This reveals lidar-scan misalignment, most likely due to sub-par lidar-inertial extrinsic calibration. This error is compounded downstream with visual-inertial calibration error which effects the accuracy of LIOSAM-derived camera poses. This is a challenge with performing non-visual camera pose estimation as every extrinsic transformation is an extra source of error which gets accumulated. The figure also highlights how the RANSAC-fitted cylinders interact with pipe reconstruction noise. It is evident that the RANSAC method biases the cylinder fit towards more circular parts of the point cloud. This could be a source of the under-estimating pattern observed in the estimated diameters across all three methods. 
\par The results show that the two NeRF reconstructions are noisy, but both have significantly less noise deviation compared to the LiDAR-inertial SLAM reconstruction as seen in Figure~\ref{fig:pvc_pipe_results1}. Despite more observed noise, the SLAM point cloud yielded the most accurate diameter estimate to within 5 mm of the expected size. The NeRF methods estimate the diameter to within 2.2-2.7 cm (5.68-7.12\%) with the iPhone data marginally outperforming the robot-sourced data. More data collection trials are necessary to understand the consistency of these error metrics. In summary, this experiment validated the ability to measure metrically-relevant scene geometry from neural-rendered scene reconstructions trained on both iPhone and robot imagery. The reconstruction quality and cylinder modeling capabilities using image-based NeRF geometry competes with the state-of-the-art technique relying on expensive LiDAR and inertial sensors.
\section{Conclusion}
In this paper, we motivate the viability of NeRF reconstruction for real-world measurement tasks. We present an experiment which involved creating two NeRF reconstructions of a 40 cm, three meter tall PVC pipe. We show that both robot and mobile phone data can be used to gather training data. We compare the fidelity of the reconstructions to a state-of-the-art technique which uses expensive 3D LiDAR and inertial sensors in combination with a SLAM algorithm (LIOSAM) to tightly fuse LiDAR frames together. We show that both NeRF-generated reconstructions are less noisy compared to the LiDAR reconstruction. Furthermore, we show that the metric accuracy of the reconstructed PVC pipe is within an acceptable range (1-2 cm) of the state-of-the-art SLAM method. 
\par The results presented in this paper offer exciting potential for neural scene representation. The ability to recover intricate scene geometry through commodity mobile phone data can accelerate the process of mapping forest environments which will provide greater insight to the state of our forests, the largest carbon sequester, and policies to help preserve this indispensable resource in the face of a changing climate. Another avenue of research is augmenting SLAM algorithms to use NeRF methods in their mapping pipeline. This can possibly mitigate the bottlenecks faced by pure visual-SLAM by introducing inference to reduce accumulated error.

\bibliography{ref}

\end{document}